# Combining SAR Simulators to Train ATR Models with Synthetic Data

Benjamin Camus[a], Julien Houssay[a], Corentin Le Barbu[a], Eric Monteux[a], Cédric Saleun[b], Christian Cochin[b]

[a] Scalian DS, 2 rue Antoine Becquerel 35700 Rennes, France
[b] DGA Maîtrise de l'Information BP 7 35998 Rennes CEDEX 9, France

## Abstract

This work aims to train Deep Learning models to perform Automatic Target Recognition (ATR) on Synthetic Aperture Radar (SAR) images. To circumvent the lack of real labelled measurements, we resort to synthetic data produced by SAR simulators. Simulation offers full control over the virtual environment, which enables us to generate large and diversified datasets at will. However, simulations are intrinsically grounded on simplifying assumptions of the real world (i.e. physical models). Thus, synthetic datasets are not as representative as real measurements. Consequently, ATR models trained on synthetic images cannot generalize well on real measurements. Our contributions to this problem are twofold: on one hand, we demonstrate and quantify the impact of the simulation paradigm on the ATR. On the other hand, we propose a new approach to tackle the ATR problem: combine two SAR simulators that are grounded on different (but complementary) paradigms to produce synthetic datasets. To this end, we use two simulators: MOCEM, which is based on a scattering centers model approach, and Salsa, which resorts on a ray tracing strategy. We train ATR models using synthetic dataset generated both by MOCEM and Salsa and our Deep Learning approach called ADASCA. We reach an accuracy of almost 88 % on the MSTAR measurements.

## 1 Introduction

Automatic Target Recognition (ATR) on SAR image is a long-standing problem that consists of automatically classifying an object of interest imaged by a radar. Numerous works of literature have demonstrated that Deep Learning is able to tackle this challenge [1, 2]. However, Deep Learning requires thousands of labelled data to train ATR models. Acquiring such diversified and complex dataset may be very expensive, or even infeasible, especially in Defense applications. Indeed, because the targets signatures vary significantly with the observation geometry, the training datasets must cover a wide angular range, with a sufficiently fine angular resolution—particularly in azimuth. Moreover, the signatures depend on the radar sensor (e.g. carrier frequency, bandwidth, thermal noise), and on the environment of the target which may also vary greatly (e.g. electromagnetic coupling, shadowing effects). Finally, the targets themselves may be highly variable because they are equipped with articulated parts (e.g. gun turret, doors) and removable components (e.g. fuel tank).

To build robust and efficient ATR models, the training datasets must encompass all these variation factors for all targets, including potential confusing objects, and systematically explore their extent as much as possible, while having ground truth information to label images. Acquiring enough real measurements to capture this combinatorial explosion is infeasible in practice. That is why the only viable alternative is to rely on SAR simulators to build synthetic datasets. With simulation, we have full control over the 3D CAD models of the targets and their virtual environment. It is then possible to produce variants at will, with different electromagnetic (EM) materials, and to run parametric production to automatically consider a large amount of observation geometries and associate ground truth information. Simulation can then capture the diversity found in real operational scenarios.

However, simulations are intrinsically grounded on simplifying assumptions of the real world (i.e. physical models). Because of that, synthetic datasets are not as representative as real measurements. Consequently, ATR models trained with synthetic images cannot generalize well on real measurements. This corresponds to the well-known Dataset Shift problem in Machine Learning, which occurs when the training and test distributions are too different [3].

The fact that the simplifications introduced by the simulation are, at least partially, responsible for the generalization issues of ATR models raises several epistemological questions. Do the modelling choices of the EM phenomenon have an impact on the ATR results? If yes, to what extent? What are the simplifying assumptions that are necessary and sufficient for ATR? In this work, we propose to study the impact of the simulation paradigm on the ATR performances to answer these questions. To this end, we use two simulators that are grounded on different assumptions.

**MOCEM is developed by Scalian DS for the DGA (the French MoD) since more than twenty years** [4, 5]. Its scattering centres model approach performs exact geometrical search to detect the canonical EM effects that dominate the target signature (i.e. diffuse, plate, dihedral, trihedral…). Once detected, the contribution of each type of effect is calculated through dedicated analytic equations.

**Salsa is a new SAR simulator developed by Scalian DS.** It relies on a ray tracing strategy to compute the target signature. The different EM paths detected are then evaluated thanks to a combination of Geometrical and Physic Optics (GO and PO). Thus, Salsa is completely agnostic to the type of radar effects being detected. They naturally emerge from the coherent summation of the EM paths.

Our contributions are twofold. On one hand, we quantify the impact of the simulation paradigm on the ATR results. Each of the numerous works of the literature uses its own simulator, CAD models, EM materials, ATR algorithms, and evaluation dataset. Comparing the SAR simulators with other things being equals is then not possible. Here, we produce two separated synthetic datasets using MOCEM on one side, and Salsa on the other side. We use the same CAD models and EM materials for the two productions. Then, using the same Deep Learning algorithm called ADASCA, we train two groups of ATR models using either the Salsa dataset or the MOCEM one. The performances of the two groups of ATR models are then measured on the MSTAR public dataset. This score directly gives us a metric to compare the representativeness of the two simulators in the ATR context, because it measures the gap between the synthetic distribution and the real one. To our knowledge, this is the first time this kind of study is done in the literature.

Our second contribution is to propose a new approach to tackle the ATR problem: combine two SAR simulators that are grounded on different paradigms to produce synthetic datasets. Our main assumption is that the simulators are complementary because their simplification errors can balance each other. In other words, we hypothesize that the union of the two synthetic distributions will better overlap the real test distribution. To this end, we train ATR model with our ADASCA approach using both the Salsa and the MOCEM datasets simultaneously. As previously, the ATR models are then evaluated on the MSTAR dataset.

The rest of the paper is organized as follows. Section 2 draws up the related works for both the ATR approaches of the literature, the existing simulation approaches and the SAR ATR dataset publicly available. Section 3 introduces the tools used in our approach (i.e. MOCEM, Salsa, and ADASCA). Section 4 details our dataset production strategy. Finally, in Section 5 we present our results.

## 2 Related works

### 2.1 ATR datasets

The MSTAR (Moving and Stationary Target Acquisition and Recognition) public dataset [6] comprises SAR measurements of fifteen different targets taken at different depression and azimuth angles by an airborne radar. The 3671 images collected at a depression angle of 17° constitute the training set whereas the 3203 images with a depression angle of 15° serve to test the models. These data concerns ten classes of vehicles (2S1, BMP2, BDRM2, BTR60, BTR70, D7, T62, T72, ZIL131 and ZSU23-4), measured almost at each azimuth degree from 0° to 360°. Three variants of vehicles are available for two classes: BMP2 and T72. The vehicles equipment and configuration (e.g. side skirts) may differ from one variant to another.

The SAMPLE (Synthetic and Measured Paired Labeled Experiment) public dataset [7] comprises pairs of real SAR measurements and simulated images. This dataset is smaller than MSTAR with only 806 synth-real measurements pairs for training (at depression angles of 14°, 15° and 16°) and 539 pairs for testing (at depression angles of 17°). Like MSTAR, SAMPLE comprises ten target classes labelled 2S1, BMP2, BTR70, M1, M2, M35, M60, M548, T72, and ZSU23-4. SAMPLE data does not provide any variant for any class. The SAMPLE dataset suffers from several drawbacks. First, the images azimuth angles range only from 10° to 80° for both training and test datasets. It particularly excludes the cardinal directions that may be the more challenging angles. Secondly, this angular sector may not be representative of the challenges met when classifying images at a full 360° extent. In addition, the SAMPLE authors have deployed considerable efforts to make the synthetic data as close as possible to real measurements, using detailed ground truth information. This very favorable scenario is unlikely to occur in an actual operational context where the measured vehicles may differ significantly from the CAD models used during training.

### 2.2 ATR approaches

Ødegaard et al. got mixed results when training an off-the-shelf deep-learning algorithm directly using simulated SAR data [8]. Using a transfer learning strategy, Malmgren Hansen et al. pre-trained an ATR classifier with a large amount of synthetic data before training the model with a smaller set of measured images [9]. The drawback of this approach is that it still requires measured data for training, and in many cases, measured images of the target of interest will not be available.

Several works focus on learning an optimal transport function to refine synthetic data by adding features peculiar to measured images. The goal is to transport the synthetic distribution on the measured one to fix the dataset-shift issue. The refined synthetic dataset can then be used to train a regular classifier. Cha et al. trained a residual network to refine synthetic data for ATR [10]. However, their classifier only achieves an accuracy of 55 % at test time.

Lewis et al. [11] and Camus et al. [12] trained a GAN to refine synthetic SAR images, with promising results of almost 95 %. However, all these refining approaches require real measurements that may be impossible to obtain. Moreover, Camus et al. demonstrated that GAN cannot refine new classes never seen during training [12].

Inkawhich et al. trained ATR classifiers with the public synthetic data of SAMPLE by combining several algorithms of the literature designed to improve the generalization of deep-learning models [13]. They evaluate all their models on the SAMPLE measured images. With several combinations of these techniques the authors found accuracy of almost 95 % on the measured images. However, the authors used the SAMPLE dataset that contains several flaws, as stated in the previous section. Camus et al. [14] have demonstrated that the results drop down to 66 % when considering a slightly less favorable scenario. Therefore, the proposed approaches may not be applicable in an actual operational scenario.

Delhommé et al. [15] used physic-based data augmentation techniques based on Attributing Scattering Centres (ASC)

parameters. By training their ATR models only on synthetic data generated by the MOCEM simulator, they reached an ATR accuracy of almost 71 % on MSTAR.

## 2.3 SAR simulations

As highlighted in [16], SAR simulators can be broadly divided into two main categories: rigorous full-wave EM simulators (e.g., MoM, FEM, FDTD), which offer high physical fidelity at the cost of significant computational resources, and high-frequency fast simulators based on asymptotic methods such as GO, PO and related hybrid ray-based or diffraction-based approaches. Only the latter category enables the generation of large-scale SAR datasets suitable for ATR development within practical computation times, while still capturing essential scattering phenomena, polarization effects, and coherent phase information. This is why the two simulators we are using here belong both to this category.

# 3 Our approach

## 3.1 Simulation

### 3.1.1 MOCEM

The main purpose of MOCEM is to convert a meshed 3D CAD model (representing the input scene to simulate) into a representation adapted to radar imagery. This representation corresponds to a scattering centers model (M3D) that lists thousands of geometrical scatterers. Each M3D is specific to an acquisition geometry (i.e. azimuth and incidence angles). An M3D is not a simple collection of isotropic scatterers localized in the 2D plan of the image. The scatterers are indeed positioned in the 3D frame of the target (see Figure 1). They also have a physical dimension and (mostly) a non-isotropic contribution. The M3D notably allows to add randomness on directive effects (e.g. plates, multibounces) to perform physic-based data augmentation in an efficient way.

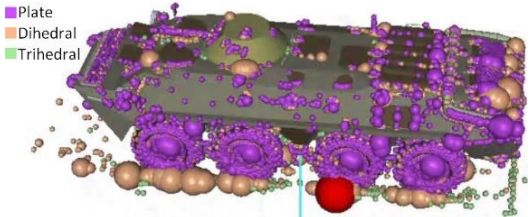

Figure 1. Example of M3D (represented as RCS spheres) over a BTR70 CAD model.

Unlike ray tracing approaches like Salsa, MOCEM only searches for the canonical radar effects that dominate the target signature (namely exact or approximate diffuse, plate, dihedral, trihedral) as well as their mirror counterparts. Their contributions are computed from analytical equations, using OP for the last facet and OG for the other ones. The effects detection is based on an original algorithm that performs an exact geometrical search through facets projection. Contrary to a ray tracing approach, this algorithm ensures that all the small elements of the CAD model are considered, and thus that no important EM contributions are ignored. Once detected, the scatterers are filtered according to geometric criteria (e.g. facets orthogonality) to speed up the calculations and to consider only the canonical effects whose contribution can be computed thanks to an analytical model. It is also possible to include additional effects that are not explicitly represented in the CAD mesh, such as edge diffraction (GTD), wire scattering, tabulated point scatterers with predefined RCS values, or local MoM computations for 1D profiles. The EM materials can be tuned to introduce surface roughness or small reliefs that are unresolved in the CAD model and compensate for insufficient faceting. For the first order interactions, MOCEM uses a backscattering model in addition to coherent and specular OG/OP effects. It simulates the non-coherent part of the signal backscattered by a rough surface. This enables to "fill" the whole visible surface of the object with EM response.

Once an M3D is produced, MOCEM rasterizes its effects into a focusing grid to form a so called "source image". The source image is an ideal SAR image without thermal noise, and with an impulse response that corresponds to a Dirac. To convert the source image into a proper radar image, MOCEM applies the transfer function of the sensor (see Figure 2), which is based on the image quality to reproduce (i.e. resolution, side-lobes, NESigma0).

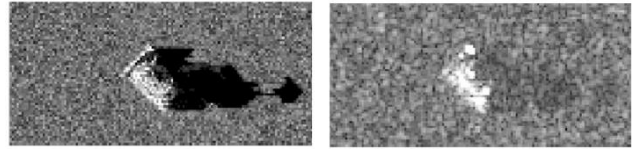

Figure 2. source image (left) vs. radar image (right) of the D7.

### 3.1.2 Salsa

Salsa was designed to rapidly compute the EM signature of meshed objects derived from CAD models, with the objective of updating this signature dynamically during a time-domain simulation for instance to produce radar raw IQ signals. The long-term goal is to enable the recalculation of environment–target couplings at each radar pulse for dynamically moving scene (like sea surface). These objectives require a careful balance between physical representativeness and computational performance.

For this reason, Salsa adopts the Shooting and Bouncing Rays (SBR) technique, implemented in a dual architecture: a CPU-based version using Intel Embree for the prototyping phase and a GPU-accelerated version using NVIDIA OptiX and CUDA for the production phase.

The EM model relies on the classical assumptions of GO to determine ray trajectories, while PO is applied on the final interaction facet to compute the complex reflected field, taking into account the properties of dielectric materials and the polarization effects.

Through these design and implementation choices, Salsa provides a fast and physically coherent framework for modeling complex manufactured targets producing multiple bounces. It is fully compatible with both monostatic and bistatic radar configurations. The ray-launch density plays a critical role in controlling the trade-off between accuracy and computing speed. A distinctive feature of Salsa is that it allows the generation of multiple types of radar products and offers a complementary and versatile approach to SAR image formation. For a given acquisition geometry, Salsa can compute the complete complex scene response, enabling holographic reconstruction through angular and frequency synthesis. It can also generate range profiles suitable for IQ data generation, which can subsequently be focused using algorithms such as Back-Projection. Both approaches can be used in bistatic configurations. Alternatively, Salsa can produce an ideal "source image" like the MOCEM approach, which can then be convolved with the radar sensor's transfer function to obtain realistic monostatic SAR images. In this study, we use source image products. Surface roughness is introduced by using a meshed rough surface for the ground. The main purpose is to sufficiently perturb the coupling effects so as to reduce their overly coherent and directive behavior that occurs with a smooth coupling surface. This introduces a more realistic diversity of target–ground interactions, better representing the variability of a rough terrain.

### 3.1.3 Simulators comparison and specific settings

*Table 1. Comparison of MOCEM and Salsa. (features in italic are not used in our study)*

|  | MOCEM | Salsa |
|---|---|---|
| Scene Input | 3D CAD models | |
| Material type | PEC and Dielectric (coefficient) | |
| Effect detection | Exact facet projection | Ray tracing |
| Maximum multi-bounce order | 3 *(+2 optional mirror reflections)* | N |
| Aspect angle for detection | Single | |
| Main EM model | GO/PO | |
| Geometrical research filter | Only canonical (tolerance-based) | No filter |
| EM Model computation | Analytic | Full computation |
| EM Backscattering and rugosity | Behavioral models | Geometrically resolved (OG/OP) |
| Additional EM effects | *Mirrors, edges, wire frames, tabulated point-scatter, specific local MoM (1D)* | NA |
| Image formation | Fast imaging | |
| Output | Source image & M3D | Source image |
| Hardware | CPU | GPU or CPU |

Although both MOCEM and Salsa are based on the same asymptotic models (GO and PO), their simulation paradigms exhibit significant differences. The Table 1 summarizes the main features of each simulator, specifically in the context of fast SAR images generation.

For this study, we configure Salsa with up to five bounces (i.e. $N = 5$) and a ray elementary surface of $1 \times 10^{-6}$ m², providing accurate modeling of the MSTAR target's EM signature. With MOCEM, we do not use the additional EM effects of Table 1 because they are not suitable for the production of large training dataset for ATR (e.g. the activation of mirror effects double the computation time, some models require detailed ground-truth information that are not available in our context). These additional effects are indeed more tailored to reproduce a specific measured image rather than to perform massive production.

Figure 3 presents a comparison between the measured BMP2 MSTAR image and those generated by the two simulation models, all displayed using a common QPM look-up table. The simulated ground clutter shows close agreement with the measured data in terms of both power level and statistical distribution. The sensor response and thermal noise are also accurately reproduced. Consequently, it is difficult to visually distinguish the simulated data from the measurements when labels are removed. However, despite their overall similarity, noticeable differences remain in the target signatures, which are expected to be partly compensated through data augmentation using ADASCA (detailed in Section 3.2). Depending on the area, MOCEM may be closer to MSTAR (yellow area), Salsa may be closer to MSTAR (red), or all three signatures may appear similar (green).

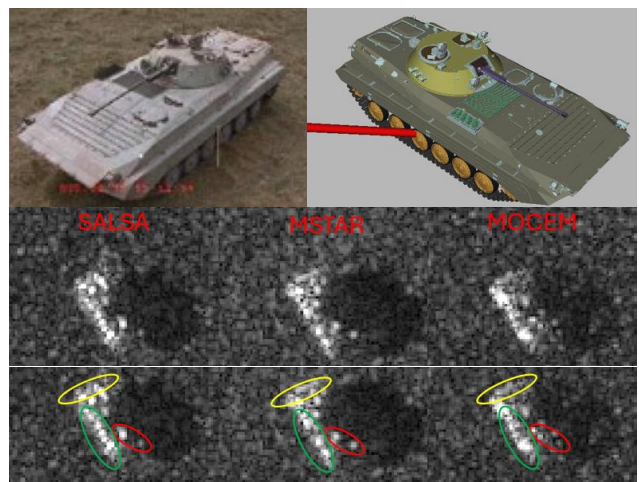

*Figure 3. Optical view of BMP2, 3D CAD model and comparison between MSTAR, Salsa and MOCEM.*

### 3.2 ADASCA

ADASCA was designed specifically to train ATR models on synthetic data [14]. It combines Deep Learning technics to improve the generalization of the ATR models with an intensive domain randomization strategy [17]. It consists of randomizing the simulated environment parameters to introduce as many variations in the synthetic data as possible. For each training epoch, we create a variant of all our synthetic dataset by randomly determining separately for each image (see Figure 4): the range and cross-range resolution, the level and distribution of the background clutter, the thermal noise of the sensor, and the target position in the images. We also randomly drop out the strongest bright points of the target.

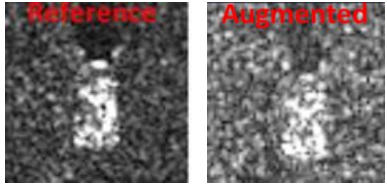

*Figure 4. Example of domain randomization.*

ADASCA applies these data augmentations directly on the source images generated by MOCEM and Salsa. It generates and applies then the sensor function to convert the source images to radar images according to the desired images quality. Since our previous work [14], we have extended ADASCA to work on oversampled source images to be more representative by taking account scatterers that are not properly centered in the pixels. The images are down sampled after the application of the sensor function to match the expected pixel size. ADASCA is implemented in Tensorflow (TF) to run in parallel on GPU and to perform all these data augmentations on batches of images directly at runtime during the classifier training. More details of our Deep Learning algorithm and architecture can be found in [14].

## 4 Datasets production

To generate our datasets with Salsa and MOCEM, we consider the 10 targets of the MSTAR dataset. Contrary to other works of literature, we do not fine-tuned our simulations (i.e. CAD models and the EM materials) to be as close as possible to the ground truth of the test data. This enables us to be closer to the context of an operational scenario.
We take off-the-shelf CAD models available on Internet. We use one CAD model per class. We manually simplify the meshes (when appropriate) to speed up the simulation time with MOCEM, and we associate generic EM materials (i.e. with well-defined reflectivity, roughness and dielectric constant) to the different facets of the model. We consider the transfer function of the MSTAR sensor (i.e. with similar range/cross-range sampling and resolution, thermal noise level, and Taylor window function). We run parametric productions for the 16°, 17° and 18° depression angles. For each depression, we generate images at every 0.5° azimuth for the full 360° range. We test our ATR models on MSTAR data that are at 15° depression angle (including the T72 and BMP2 variants). Thus, we do not consider the exact same azimuth and incidence angles for training and test.

## 5 Results

When the ATR models are trained on the MOCEM dataset, they reach an accuracy of 80.58 % (reminds that CAD models and materials are not fine-tuned to fit the test images). This represents an increase of 4 % compared to our previous work [59]. This comes from the use of oversampled source images in ADASCA. With the ATR models trained with the Salsa dataset, we measured an accuracy of 86.35 %. Thus, we found a gap of 5.77 % between the two simulators. We conclude that there is a significant impact of the simulation paradigm on the ATR results.

The combination of the two simulators further increases this result by 1.6 % to reach a total of 87.91 %. Considering that the Salsa data alone already have an accuracy of 86 % and that there is only 14 % improvement that remains possible to reach 100 %, we consider this 1.6 % improvement to be significant (even more so we do not know if an accuracy of 100 % is even possible in an operational context). The confusion matrix is shown in Figure 5. We observe that the ATR models reach an accuracy of 99 % for the D7. For six classes (2S1, BMP2, BRDM2, BTR60, BTR70, and D7) we found an accuracy close to or higher than 90 %. The ZIL131 and ZSU23-4 classes are at about 84 %. The lowest accuracies are for the T62 and T72 classes (81.7 % and 76.1 % respectively). However, we notice an important confusion between these two vehicles (20.4 % of the T72 images are confused with the T62). This is not surprising because the two vehicles are very similar (both are main battle tanks with comparable dimensions). The errors made by the ATR models are then consistent. If we consider a common meta-class "main battle tanks" for the T62 and the T72, our ATR models reach an accuracy of 92.14 %.

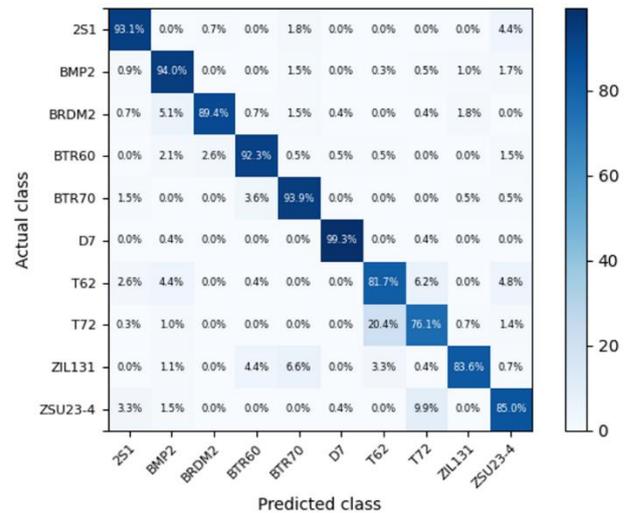

*Figure 5. confusion matrix of the ATR models trained on both MOCEM and Salsa data.*

## 6 Conclusion

In this work, we compared and combined two SAR simulators, MOCEM and Salsa, to train ATR models. This enables us to study and improve the impact of the physical modeling paradigm of simulation on the ATR. We have demonstrated that ATR models trained simultaneously on synthetic data generated by MOCEM and Salsa (and only on synthetic data) can reach an accuracy of almost 88 % on the real MSTAR data measurements. This represents a significant increase compared to other works of literature. It is important to note that to obtain these results, we did not fine-tune our simulations to unrealistically match the ground truth of the test measurements. Also, this mean score includes angles where classification might be very difficult even impossible, especially on specular directions.

The two simulators we studied are not radically different from each other. Both are based on an OG/OP mixture with a fast image rendering strategy. This enables us to measure the impact of a restricted set of modelling assumptions (i.e. the detection and evaluation of the EM effects) on the ATR performance. Interestingly, we found that even these restricted differences led to a significant gap of 5.77 % on the ATR precision. In future work, we plan to progressively extend Salsa to change other modelling choices. In this way, we will be able to measure the impact of these choices on the ATR. For instance, we will add the capacity to generate raw IQ signals and create a SAR image using a focusing algorithm instead of fast images to compare the two rendering strategies. This approach allows a finer simulation of the radar image formation process compared to fast-imaging methods. It accounts for effects inherently linked to SAR azimuthal/angular integration and to the focusing algorithm, such as phase evolution, gradual shadowing, defocusing artifacts, and other related phenomena. Also, in addition to the OG/OP model, we plan to enhance physical modeling by using for instance the theory of diffraction.

In our experiments, we found better results with Salsa than MOCEM. However, it must be noted that the MOCEM paradigm enables some data augmentations that cannot be done by Salsa. Indeed, we can efficiently perform physic-based data augmentation directly on the M3D produced by MOCEM. For instance, we can add noise on the 3D position of some scattering centers or randomly add or remove some effects. It has been shown in the literature that similar data augmentations can improve significantly the ATR results [15]. We did not use these techniques here to be able to compare the two simulators other things being equals. In future work, we plan to measure the impact of the M3D augmentations on the MOCEM results.

We found that combining the two simulators significantly improves the ATR precision of 1.6 %. This demonstrates that Salsa and MOCEM are complementary because the union of their synthetic dataset distributions better overlaps the real one. However, the increment remains limited. In future work, we plan to refine the combination by forming hybrid images composed of EM effects from both MOCEM and Salsa. This should increase the representativeness of the individual images.

It is important to note that we obtained similar results with a single-blind study (i.e. the team that produced the synthetic data and trained the ATR models have never seen the test data) on another real measurements dataset, with different targets, radar sensor, acquisition geometry. This strengthens our confidence in the reproducibility of our approach. Real measurements of the targets appear more useful to adjust simulation parameters (size, localization of equipment, levels on EM materials) and/or to validate the ATR models, than to use it in the AI learning process.

# 7  Acknowledgement

This work has been funded by the AID (Agence Innovation Defense) and the DGA (French MoD) in the context of the TACOS project. Thanks to them.

# 8  Literature

[1] Y. Li, et al. "DeepSAR-Net: Deep convolutional neural networks for SAR target recognition," 2017 IEEE ICBDA, 2017, pp. 740-743

[2] D. Morgan. "Deep convolutional neural networks for ATR from SAR imagery." ASARI XXII. Vol. 9475. SPIE, 2015.

[3] J. Quionero-Candela, M. Sugiyama, A. Schwaighofer, and N.D. Lawrence. 2009. Dataset Shift in Machine Learning. The MIT Press.

[4] C. Cochin, P. Pouliguen, B. Delahaye, D. Le Hellard, P. Gosselin, and F. Aubineau, "Mocem - an 'all in one' tool to simulate sar image," pp. 1 – 4, 07 2008

[5] C. Cochin, J.-C. Louvigne, R. Fabbri, C. Le Barbu, A. O Knapskog, and N. Ødegaard, "Radar simulation of ship at sea using mocem v4 and comparison to acquisitions," *International Radar Conference* 2014.

[6] T. Ross, S. Worrell, V. Velten, et al. , "Standard SAR ATR evaluation experiments using the MSTAR public release data set," Proc. SPIE 3370, ASARI V, 1998

[7] B. Lewis, et al. "A SAR dataset for ATR development: the Synthetic and Measured Paired Labeled Experiment (SAMPLE)." ASARI XXVI. Vol. 10987. SPIE, 2019.

[8] N. Ødegaard, A. O. Knapskog, C. Cochin and J. Louvigne, "Classification of ships using real and simulated data in a convolutional neural network," 2016 IEEE RadarConf, 2016.

[9] D. Malmgren-Hansen et al., "Improving SAR Automatic Target Recognition Models With Transfer Learning From Simulated Data," in IEEE GRSL, vol. 14, no. 9, . 2017.

[10] M. Cha, et al., "Improving Sar Automatic Target Recognition Using Simulated Images Under Deep Residual Refinements," IEEE ICASSP, 2018.

[11] B. Lewis, J. Liu, A. Wong, "Generative adversarial networks for SAR image realism," SPIE, ASARI XXV, 2018.

[12] B. Camus, E. Monteux and M. Vermet. Refining Simulated SAR images with conditional GAN to train ATR Algorithms. In Proc. CAID, 2020.

[13] N. Inkawhich et al., "Bridging a Gap in SAR-ATR: Training on Fully Synthetic and Testing on Measured Data," in IEEE JSTARS, vol.14, pp.2942-2955, 2021.

[14] Camus, B., Barbu, C. L., Monteux, E. (2022). Robust SAR ATR on MSTAR with Deep Learning Models trained on Full Synthetic MOCEM data. In Proc. CAID'22.

[15] E. Delhommé, H. Remusati, C. Lesueur, J. Petit-Frère Physics-inspired data augmentation for SAR ATR: a new approach to tackle the synthetic-to-measured Domain Gap. In Proc. CAID 2025.

[16] Woollard, M., et.al., M.A. SARCASTIC v2.0—High-Performance SAR Simulation for Next-Generation ATR Systems. Remote Sens. 2022, 14, 2561.

[17] Tobin et al. Domain randomization for transfering deep neural networks from simulation to the real world. IEEE IROS 2017